\documentclass[11pt,a4paper]{article}
\usepackage[hyperref]{eacl2021}
\usepackage{times}
\usepackage{latexsym}
\usepackage{linguex}
\usepackage[utf8]{inputenc}
\usepackage[T1]{fontenc}
\usepackage{expex}
\usepackage{tabto}
\usepackage{gb4e}
\usepackage{footnote}
\noautomath
\usepackage{graphicx}
\usepackage{epstopdf}
\usepackage{natbib}
\usepackage{tipa}
\usepackage{natbib}
\usepackage{tabularx}
\usepackage{soul}
\usepackage{makecell}

\usepackage{caption}
\usepackage{subcaption}
\usepackage{subcaption}
\usepackage{minipage-marginpar} 
\usepackage{float}
\usepackage{epstopdf}
\usepackage{appendix}
\usepackage{array,booktabs,makecell}
\usepackage{adjustbox}
\usepackage{microtype}
\usepackage{hyperref}
\usepackage{xstring}
\usepackage{multirow}
\usepackage{booktabs}
\usepackage{tablefootnote}

\usepackage{microtype}

\aclfinalcopy

\title{Translating the Unseen? Yor\`{u}b\'{a}$\rightarrow$English MT in Low-Resource, Morphologically-Unmarked Settings}

  \author{Ife Adebara$^1,^2$ ~~~~~~~~ Muhammad Abdul-Mageed$^1,^2$   ~~~~~~~~ Miikka Silfverberg$^1$  \\
\normalsize 
    Department of Linguistics$^1$ \\
    Natural Language Processing Lab$^2$  \\
  \normalsize The University of British Columbia\\
  \texttt{ \small \{ife.adebara,muhammad.mageed,miikka.silfverberg\}@ubc.ca}
  }
\date{}

\begin{document}
\maketitle
\begin{abstract}
Translating between languages where certain features are marked morphologically in one but absent or marked contextually in the other is an important test case for machine translation. When translating into English which marks (in)definiteness morphologically, from Y{o}r\`{u}b\'{a} which uses bare nouns but marks these features contextually, ambiguities  arise. In this work, we perform fine-grained analysis on how an SMT system compares with two NMT systems (BiLSTM and Transformer) when translating \textit{bare nouns} in Y{o}r\`{u}b\'{a} into English. We investigate how the systems what extent they identify BNs, correctly translate them, and compare with human translation patterns. We also analyze the type of errors each model makes and provide a linguistic description of these errors. We glean insights for evaluating model performance in low-resource settings. In translating bare nouns, our results show the transformer model outperforms the SMT and BiLSTM models for $4$ categories, the BiLSTM outperforms the SMT model for $3$ categories while the SMT outperforms the NMT models for $1$ category. 
\end{abstract}


\section{Introduction}\label{sec:intro}
Languages differ with regard to how grammatical information such as ``case" and ``number" are expressed. In some languages, this information is overtly marked using morphological or syntactic means, whereas in others it has to be inferred from context. This asymmetry of information representation poses an important problem for machine translation \cite{mitkov1999introduction,hardmeier2012discourse}. Several phenomena where asymmetry arises have been identified as challenging problems for machine translation. These include: pronoun translation and coreference \cite{guillou2019findings}, politeness \cite{sennrich2016controlling}, lexical cohesion \cite{carpuat2009one}, and lexical disambiguation \cite{gonzales2017improving}. 

Asymmetry of information representation presents an interesting test case for MT systems because it can shed light on their true linguistic ability \cite{voita2018context, bawden2017evaluating}. It is, therefore, important to use evaluation measures which can capture this aspect of the translation task. However, the most popular evaluation metric for machine translation, the BLEU score \cite{papineni-etal-2002-bleu}, is a coarse metric which can often hide these fine-grained morphological and semantic distinctions. In fact, a high BLEU score is no guarantee of improved translation quality and BLEU, being based on precision on short ngrams, may be poorly suited for measuring the coherence and grammaticality of a sentence. 

In this paper we investigate the performance of Y{o}r\`{u}b\'{a} \footnote{Y{o}r\`{u}b\'{a} is a tone language that belongs to the Yoruboid group of the Kwa branch of the Niger-Congo language family, which is spoken by over 40 million in Nigeria. Y{o}r\`{u}b\'{a} is spoken primarily in western Nigeria and eastern Benin, with communities in Sierra Leone and Liberia, and expatriate communities throughout Africa, Europe, and the Americas.} to English Machine Translation. We specifically evaluate performance on translating \textit{Bare nouns} (BNs). BNs ~\cite{cheng1999bare, krifka2003bare, chierchia1998reference, larson1985bare, carlson1989semantic} are nouns without an overt determiner or quantifier. For instance "houses" in \textit{Houses are expensive in New York} is a BN. Whereas English accounts for only plural BNs, BNs in Y{o}r\`{u}b\'{a} are number neutral and can also be definite or indefinite depending on the context. Consider the following example: 

\pex<withparts> 
\footnotesize 
\begingl 
\gla    B\`{a}b\'{a} ra i\textsubdot{s}u //
\glb \textsc{father}  \textsc{buy} \textsc{yam} //
\endgl

\a \textit{`Father bought a yam.'} \quad\textbf{(Indefinite Singular)}
\a \textit{`Father bought some yams.'}\quad\textbf{(Indefinite Plural)}
\a \textit{`Father bought the yam.'}\quad\textbf{(Definite Singular)}
\a \textit{`Father bought the yams.'}\quad\textbf{(Definite Plural)}
\xe

In Example (1), the BN \textit{i\textsubdot{s}u} {\it yam} can be translated into English in four ways as: the indefinite singular \textit{a yam}, an indefinite plural \textit{some yams}, a definite singular \textit{the yam}, and a definite plural \textit{the yams}. This poses a challenge akin to anaphora resolution, as the correct translation of Y{o}r\`{u}b\'{a} BNs can only be determined by examining the context in which the BN occurs in the source text. The context can span one or more preceding or current words, phrases, clauses or sentences. It can also include world knowledge.

Our study provides a fine grained analysis that sheds light on issues in MT that are not often discussed in main stream research. We turn away from the current research trend in massively multilingual translation systems, to this largely underexplored fine-grained aspect of the MT problem. We investigate how SMT and NMT systems compare in general, but also specifically w.r.t. translation of BNs. 

Although NMT has recently been reported to outperform SMT even in low-resource settings, these findings were reported for systems translating between somewhat similar languages (e.g., languages that belong to the same language family, have overlapping vocabulary and similar script) \cite{junczys2016neural, conneau2017word, lample2017unsupervised, wang2019learning} such as English $\rightarrow$ French, German $\rightarrow$ French, German $\rightarrow$ Italian; languages for which large corpora exists. In this work, we collect a new dataset and use it to test whether the same NMT advantage persists by exploring two \textit{typologically dissimilar} languages belonging to different language families: Y{o}r\`{u}b\'{a} and English. This work bears significance not only to Y{o}r\`{u}b\'{a} but to analytic languages, low-resource languages, and languages that have BNs. 

Our contributions are as follows: \textbf{(1)} We align a new dataset for the Y{o}r\`{u}b\'{a} and English low-resource setting. \textbf{(2)} We use our dataset to develop statistical and NMT Y{o}r\`{u}b\'{a} $\rightarrow$  English models. \textbf{(3)} We study the linguistic ability of our models in disambiguating BNs. 

The rest of the paper is organized as follows: Section \ref{sec:problem} is a description of disambiguation patterns of Y{o}r\`{u}b\'{a} BNs. We discuss related work in Section \ref{sec:lit}. Section \ref{sec:datas} presents the datasets, data collection process, and preprocessing. In Section \ref{sec:mactrans}, we present our methods. We present results of our models and an evaluation of BN disambiguation in Section \ref{sec:results}. We conclude in Section \ref{sec:conc}.

\section{Disambiguating Y{o}r\`{u}b\'{a} BNs }\label{sec:problem}
A number of contextual variables are important when disambiguating BNs in Y{o}r\`{u}b\'{a}. These include the so-called \textit{familiarity} and \textit{uniqueness} conditions \cite{roberts2003uniqueness,russell1905denoting, abbott2006definite}, as well as the \textit{category of the verb} that occurs in the environment of the BN. Familiarity refers to information which is already known from the previous textual context either explicitly or through inference. In Y{o}r\`{u}b\'{a}, a definite interpretation is permitted when the existence of the entity referred to has been established in the discourse, an indefinite occurs otherwise. In example (2)\footnote{1st Person Singular (1SG), Perfective (PERF)}, the first mention of book is indefinite, the second is definite.

\pex<withparts> 
\footnotesize 
\begingl 
\hspace{\fill}
\gla Mo ra \textbf{\`{i}w\'{e}} f\'{u}n K\textsubdot{\'{o}}l\'{a}. L'\textsubdot{\'{o}}j\textsubdot{\'{o}} kej\`{i}, K\textsubdot{\'{o}}l\'{a} ti so \textbf{\`{i}w\'{e}} n\`{u}  //
\hspace{\fill}
\glb \textsc{1SG} \textsc{buy} \textsc{book}  \textsc{for} \textsc{K\textsubdot{\'{o}}l\'{a}}. \textsc{In day} \textsc{second} \textsc{K\textsubdot{\'{o}}l\'{a}} \textsc{PERF}  \textsc{throw} \textsc{book}  \textsc{away} \footnote{1SG = First person singular; PERF = Perfective aspect} //
\hspace{\fill}
\glft `I bought a book for \textsc{K\textsubdot{\'{o}}l\'{a}}. By the second day, K\textsubdot{\'{o}}l\'{a} had lost the book.'  //
\hspace{\fill}
\endgl
\xe

The uniqueness condition claims that there exists one and only one entity that meets the descriptive content of the BN as in example (3)\footnote{Habitual (HAB), Progressive (PROG)}. This entails that the BN will be interpreted as definite and singular.  

\pex<withparts>
\footnotesize
\begingl 
\hspace{\fill}
 \gla \textbf{O\`{o}r\`{u}n} m\'{a}a \'{n} r\`{a}n n\'{i} \textbf{\textsubdot{\`{o}}s\'{a}n} //
 \hspace{\fill}
\glb \textsc{sun}  \textsc{Hab}  \textsc{Prog}  \textsc{shine} \textsc{in} \textsc{afternoon} //
\hspace{\fill}
\glft `The sun shines in the afternoon.' //
\hspace{\fill}
\endgl
\xe

The category of verb is also an important contextual element in correctly translating BNs. \textit{Stative} verbs (verbs that describe the state of being or situation such as {\it to own} and {\it to feel}) introduce the \textit{Generic} description where a noun phrase is used to refer to a whole class as in example (4). \textit{Eventive} verbs (verbs that describe events such as {\it to break} and {\it to appear}) introduce all other disambiguation patterns as in example (1). 

\pex<withparts> 
\small
\begingl 
\gla    \textsubdot{O}m\textsubdot{o}  f\textsubdot{\`{e}}r\'{a}n \textbf{aj\'{a}}  \quad \textbf{(generic)} \textbf{(STATIVE)} //
\glb \textsc{child}  \textsc{love} \textsc{dog} //
\glft \textit{`Children love dogs'} //
\endgl
\xe

\section{Related work}\label{sec:lit}

Traditional evaluation methods like BLEU are inadequate for evaluation of fine-grained discourse phenomena in MT, and various approaches have been explored to alleviate this problem. \newcite{guillou2016protest} present the PROTEST pronoun translation test suite. The test suite contains examples of pronoun translation which are known to be challenging for MT systems. \newcite{isabelle2017challenge} present another challenge set for MT. This set consists of a number of human annotated sentences which are designed to probe a system’s capacity to handle various linguistic phenomena (like EXAMPLES). Using the challenge set, \newcite{isabelle2017challenge} present a comparison of SMT and NMT systems for English-French MT which provides a fine-grained exploration of the strengths of NMT, as well as insight into linguistic phenomena that typically present difficulties for NMT models.

\newcite{sennrich2016grammatical} assess the grammaticality of the output of a character-level NMT system by evaluating the MT model's capacity to correctly rank contrastive pairs of pre-existing translations, one of which is correct and the other one incorrect. This approach has also been applied to lexical disambiguation of English-German MT  \cite{gonzales2017improving}. 
\newcite{bentivogli2016neural} explore automatic detection and classification of translation errors based on manual post-edits of MT output both for SMT and NMT systems. They use a classification that evaluated outputs for morphological, lexical, and word order errors which was a simplification of those used in Hjerson \cite{popovic2011hjerson}. Hjerson detects word level error classes: morphological errors, re-ordering errors, missing words, extra words and lexical errors.

Recently, neural models have been evaluated for syntactic competence. For instance, \newcite{linzen2016assessing} probe the ability of LSTM models to learn English subject-verb agreement. When provided explicit supervision, LSTMs were able to learn to perform the verb number agreement task in most cases, although their error rate increased on particularly difficult sentences. NMT systems have also been evaluated for morphological competence while translating from English to a morphologically rich language \cite{burlot2017evaluating}. 

Certain linguistic phenomena have also been tested across language families. For instance, for four language families: Slavic, Germanic, Finno-Ugric and Romance, the best NMT system outperformed the best phrase-based SMT system for all language directions to English \cite{toral2017multifaceted}. The NMT systems produced fluent and more accurate inflections and word order but performed poorly when translating long sentences \cite{vanmassenhove2019lost, bentivogli2016neural}. Morphologically rich languages have also been shown to have more fluent outputs \cite{klubivcka2017fine, toral2017multifaceted, popovic2018language}. These studies have used metrics such as BLEU \cite{papineni2002bleu}, HTER \cite{snover2006study}, TTR \cite{templin1957certain}, YULE \cite{yule2014statistical}  to evaluate the output of MT models for fluency and adequacy. 
\section{Dataset}\label{sec:datas}

We use the Y{o}r\`{u}b\'{a} Bible \textit{B\'{i}b\'{e}l\`{i} M\'{i}m\'{o} n\'{i} \`{E}d\`{e} Y{o}r\`{u}b\'{a}  \`{O}de-\`{O}n\'{i}} (BMEYO) and the New International Version (NIV) English Bible.

\subsection{Bible Data}
We crawled the BMEYO Y{o}r\`{u}b\'{a} Bible from the public website Biblica.\footnote{\url{https://www.Bible.com/versions/911-ycb-bibeli-mim-ni-ede-yoruba-ode-oni}.} This version is the modern translation of the Y{o}r\`{u}b\'{a} Bible and is the closest equivalent to the English NIV Bible, according to Biblica. We thus use the NIV English translation with the Bible to create our parallel data. We organize the Bible according to their verses. In Table~\ref{tab:Biblestats}, we show an example verse from our Bible dataset. 

\begin{table}[h]
\small
\begin{center}
\footnotesize 
\begin{tabular}{p{1.5 cm} p{5 cm}} 
\hline
 \thead{\textbf{Data}} & \thead{\textbf{Scripture} }
 \\ \hline  \hline
\textbf{BMEYO} & {\`{I}gb\`{a} l\'{a}ti pa \`{a}ti \`{i}gb\`{a} l\'{a}ti
 m\'{u} l\'{a}rad\'{a} \`{i}gb\`{a} \`{a}ti w\'{o}
 lul\textsubdot{\`{e}} \`{a}ti \`{i}gb\`{a} l\'{a}ti k\textsubdot{\`{o}}}
 \\ \hline
\textbf{NIV} & {a time to kill and a time to heal, a time to tear down and a time to build.}
\\ \hline
\end{tabular}
\caption{A description of Ecclesiastes Ch. $3$, Verse $3$ for the BMEYO. \textbf{(NIV English Bible translation)}}\label{tab:Biblestats}
\end{center}
\end{table}

\begin{table}[ht]
\small
\begin{center}
\begin{tabular}{c c c c} 
\hline
 \thead{\textbf{Data}} & \thead{\textbf{\#TOK}} & \thead{\textbf{\#SENT}} & \thead{\textbf{\#TTR}}  \\ 
 \hline  \hline
 \textbf{BMEYO} & 793,870  & 38,149 & 25.5  \\  \hline
\end{tabular}
\end{center}

\caption{A statistical description of the Y{o}r\`{u}b\'{a} (source) data. \#TOK refers to number of tokens, \#SENT is number of sentences, and TTR is type token-ratio.}\label{tab:stats}
\end{table}

\subsection{Data Preprocessing}
The Y{o}r\`{u}b\'{a} Bible is based on old Bible manuscripts. A total of $16$ verses which were included in early English Bible versions like the King James Bible but which were omitted from later versions
are still part of the Y{o}r\`{u}b\'{a} Bible. Therefore, we start  
data preprocessing by adding these missing verses into our English NIV Bible to make it equivalent with the Y{o}r\`{u}b\'{a} text. Those verses were footnotes in the English NIV. In addition, the book of Third John has $15$ verses in the Y{o}r\`{u}b\'{a} Bible but $14$ verses in the English NIV. The $15$\textsuperscript{th} verse in the Y{o}r\`{u}b\'{a} is a part of $14$\textsuperscript{th} verse in the English Bible. As a result, we combine verse $15$ into verse $14$, as it is in the English NIV. The aligned dataset can be found on GitHub at \textbf{\url{https://github.com/UBC-NLP/africaNLP2021}}. 

Next, we tokenized the English data using \texttt{SpaCy}\footnote{\url{https://spacy.io/}}. \texttt{SpaCy} currently does not provide a tokenization package for Y{o}r\`{u}b\'{a}, so we used the whitespace tokenizer for all the Y{o}r\`{u}b\'{a} data. We use python scripts to ensure the punctuations is appropriately tokenized. Next, we convert all words to lowercase in order to alleviate data sparsity. 

We also split words using Byte Pair Encoding (BPE) \cite{sennrich2015neural}. In low-resource settings, large vocabularies result in the representation of low-frequency (sub)words as BPE units which affects the ability to learn good high-dimensional representations \cite{barone2017deep}. Thus, we choose smaller merge operations and varied the number of merge operations from $10,000$ to $30,000$. Finally, we split the dataset into training, validation, and test sets using an $80\%$-$10\%$-$10\%$ standard split.

\section{Methods}\label{sec:mactrans}
We train 3 sentence level models: SMT, BiLSTM, and Transformer models to translate from Y{o}r\`{u}b\'{a} to English. We choose English as our target language because we are interested in analyzing how ambiguous BNs in Y{o}r\`{u}b\'{a} are translated into English where (i.e., in English) nouns are typically marked both for number and determinacy. The hyperparameters and training procedure are described in the next subsections.

\subsection{SMT}
We use the \texttt{Moses}\footnote{\url{https://github.com/moses-smt}} statistical translation system for our SMT model. We apply tokenization, true casing, and perform word alignment on the parallel data using GIZA++ \cite{och2003systematic}. The word alignments were used to extract phrase-paired translations and calculate probability estimates \cite{koehn2007moses}. We used KENLM \cite{heafield2011kenlm} to train and query a LM for English. KENLM is a library implemented for efficient language model queries, reducing both time and memory costs, and is integrated into \texttt{Moses}. The decoder uses this LM to ensure a fluent output of the target language, in our case, English. We used the validation sets of our parallel data for the final tuning process just before we perform the blind testing.

\subsection{BiLSTM}

We use a Sequence to Sequence (Seq2Seq) BiLSTM with attention model. 
Our best BiLSTM model has an embedding layer with $1,024$ dimensions, and $2$ encoder and decoder layers each.\footnote{Model architecture and hyperparameter values are identified on validation data using values listed in Table~\ref{tab:hyp}.} We use the Adam optimizer with a learning rate of $5$e$-4$ and a batch size of $32$. For regularization, we use a dropout of $0.2$.

\begin{table}[H]
\small
\begin{center}
\begin{tabular}{p{4cm}p{2cm}} 
\hline
\thead{\textbf{Hyperparameter}} & \thead{\textbf{Values}} \\
\hline  \hline

encoder layers & 2, 3, 4, 6, 8 \\ 
 \hline
decoder layers & 2, 3, 4, 6, 8 \\ 
 \hline
embedding dimension & 256, 512, 1024. \\
 \hline
batch sizes & 32, 64, 128 \\
\hline
number of tokens & 4000, 4096 \\
\hline
dropout & 0.2, 0.3, 0.4, 0.6, 0.8 \\
\hline

 \end{tabular}
 \end{center}
 \caption{Hyperparameter settings for tuning BiLSTM and Transformer models.}\label{tab:hyp}
 \end{table}

\subsection{Transformer}

For the transformer model, we use $5$ layers with $8$ attention heads in both encoder and decoder. We use embedding dimension with $1,024$ units. We express our batch size in number of tokens, and set it to $4,096$. Detailed hyperparameter settings is available in Table~\ref{tab:transhyp}

\begin{table}[H]
\small
\begin{center}
\begin{tabular}{p{4cm}p{2cm}} 
\hline
\thead{\textbf{Hyperparameter}} & \thead{\textbf{Values}} \\
\hline  \hline
    
adam-betas & (0.9,0.98) \\
\hline
clip-norm & 0.0 \\
\hline
learning rate & 5e-4 \\
\hline
learning rate scheduler & inverse square root \\
\hline
warmup-updates & 4000 \\
\hline
dropout & 0.3 \\
\hline
weight-decay & 0.0001 \\
\hline
criterion & label smoothed cross entropy \\
\hline 
label-smoothing & 0.5 \\
\hline
encoder-layerdrop & 0.2  \\
\hline
decoder-layerdrop & 0.2 \\
\hline

 \end{tabular}
 \end{center}
 \caption{Hyperparameters for Transformer model}\label{tab:transhyp}
 \end{table}

\subsection{Hyperparameters, Vocab. \& Training}
\textbf{Hyperparameters.} We experimented with different hyperparameter values to ensure optimization of our models. Since the size of data is small, we used fewer layers, and smaller batch sizes as referenced in literature \cite{sennrich2019revisiting, nguyen2017transfer}. A full range of hyperparameters and values are in Table~\ref{tab:hyp}.

\noindent\textbf{Vocabulary Size.}
We varied the number of BPE merge operations from $10,000$ to $30,000$. Our optimal models used $10$K merge operations. 

\noindent\textbf{Training.} We train the model with fairseq toolkit for $7$ days, on $1$ GPU, and choose best epoch on our development set, reporting performance on TEST. 
\section{Evaluation}\label{sec:results}
In this section, we first evaluate using BLEU. We provide details of this part of the evaluation in Section \ref{subsec:MP}. Next, we provide details on our approach to evaluating BNs in Section \ref{subsec:MP-BNs}.  
\subsection{Model Performance}\label{subsec:MP}
We evaluate the output of our models using BLEU \cite{papineni2002bleu}. We use BLEU score because it is the most commonly used metric for MT evaluation. Table \ref{tab:res} shows our n-gram precision figures from 1 to 4-grams and our overall BLEU score. Our transformer model outperforms the BiLSTM and SMT models respectively. 

We give examples of the output from each model and the gold data in Table~\ref{tab:comp}. 

\begin{table}[H]
\small

        \begin{tabular}
        {p{1.1cm} p{1.1cm} p{1.1cm} p{1.1cm} p{1.1cm}} 
        \multicolumn{5}{c}{SMT}\\
        \hline
        \textbf{1-gram} & \textbf{2-gram} & \textbf{3-gram} & \textbf{4-gram}  & \textbf{BLEU} \\
        \hline \hline
        $62.00$ & $47.97$ & $38.68$ & $31.92$ &  $33.01$\tablefootnote{BLEU = $33.01$~~ $63.0$/$38.1$/$26.2$/$18.9$ (BP = $1.000$ ratio = $1.023$ hyp$\_$len = $83606$ ref$\_$len = $81700$).}\\   \hline
    \hline
    \multicolumn{5}{c}{$ $} \\
    \multicolumn{5}{c}{BiLSTM}\\
    \hline
           \textbf{1-gram} & \textbf{2-gram} & \textbf{3-gram} & \textbf{4-gram} & \textbf{BLEU} \\ \hline \hline
         $62.97$ & $49.21$ & $40.08$ & $33.43$ &  $33.44$\tablefootnote{BLEU = $33.44$~~ $62.9/38.4/26.6/19.5$ (BP = $1.000$ ratio = $1.005$ hyp$\_$len = $82133$ ref$\_$len = $81700$).} \\  \hline
             \multicolumn{5}{c}{$ $} \\
    \multicolumn{5}{c}{Transformer}\\
    \hline
        \textbf{1-gram} & \textbf{2-gram} & \textbf{3-gram} & \textbf{4-gram}  & \textbf{BLEU} \\ \hline \hline
        $\bf66.27$ & $\bf53.42$ & $\bf44.45$ & $\bf37.68$ & $\bf37.93$ \tablefootnote{BLEU = $37.93$~~ $66.9/43.5/31.2/23.3$ (BP = $0.994$ ratio = $0.994$ hyp$\_$len = $81231$ ref$\_$len = $81700$).} \\ \hline 
       \end{tabular}
    \caption{Model performance at 1-4 grams with final column showing BLEU score as given by the sacrebleu ~\cite{post-2018-call}. }\label{tab:res}
\end{table}

\begin{table}[H]
\small
\begin{center}
\begin{tabular}{p{1.5cm}p{1.4cm}p{1.4cm}p{2cm}} 
\hline
\thead{\textbf{Gold Data}} & \thead{\textbf{SMT}} & \thead{\textbf{BiLSTM}} & \thead{\textbf{Transformer}} \\ 
\hline  \hline

as in the days when you came out of egypt, i will show them my wonders. & like the day she came up out of egypt, i will show wonders known him.  & as soon as he came out of egypt, i will show him all the wonders he has done. & like the day he came up out of egypt, i will show him the wonders. \\ 
 \hline
rescue me from the mouth of the lions; save me from the horns of the wild oxen. & the lord . deliver my life from the lions; rescue me from the horns of the wild ox. & deliver my life from the lion's hands ; deliver me from the hand of the wild ox. & rescue my life from the lion; rescue me from the horns of the wild ox. \\ \hline

 so now the lord has put a lying spirit in the mouths of all these prophets of yours. the lord has decreed disaster for you. & so the lord has put a lying spirit in the mouths of all your prophets. the lord has decreed disaster for you. &  so the lord has put a lying spirit in the mouths of all these prophets of yours. the lord has decreed disaster for you. & so the lord has put a lying spirit in the mouths of all these prophets of yours. the lord has decreed disaster for you.\\  \hline
\end{tabular}
\end{center}
\caption{A comparison of the gold data with the output of our models.}\label{tab:comp}
\end{table}

\subsection{Preparing a Gold Standard for Bare Noun Disambiguation}\label{subsec:MP-BNs}
We group English translations of Y{o}r\`{u}b\'{a} BNs into $5$ categories described in Table \ref{tab:categories}: \textit{generic}, \textit{indefinite singular}, \textit{indefinite plural}, \textit{definite singular} and \textit{definite plural}. We do this  both for the gold standard target data and for the output of our MT systems, and compare the distribution of categories in the MT output to the category distribution in the English gold standard data. The annotation was performed by a linguist who is also a native speaker of Y{o}r\`{u}b\'{a}. 

\begin{table}[H]
\small
\begin{center}
\begin{tabular}{p{2.5cm} p{4cm}} \hline
\thead{\textbf{Category}} & \thead{\textbf{Description}} \\
 \hline \hline
 Generic &  noun refers to a kind or class of individuals.   \\ 
 Indefinite Singular & noun refers to a single non-specific object \\
 Indefinite Plural & noun refers to multiple non-specific objects  \\
 Definite Singular & noun refers to a single specific object \\ 
 Definite Plural & noun refers to multiple specific objects \\
 \hline

\end{tabular}
\end{center}
\caption{Noun Categories when translating from Y{o}r\`{u}b\'{a} to English }\label{tab:categories}
\end{table}

Determining the correct category for the translation of a BN requires us to control for the following contextual information: \textbf{type of verb} (STATIVE versus EVENTIVE); \textbf{the discourse context} (e.g. the preceding sentence, the familiarity constraint); and \textbf{real-world knowledge} (which may trigger the uniqueness constraint). E.g. Y{o}r\`{u}b\'{a} words (\textsubdot{o}ba  \textit{king}, \textsubdot{o}l\textsubdot{\`{o}}run  \textit{god}) should always be translated into definite singulars because Y{o}r\`{u}b\'{a} speakers commonly use them to refer to specific entities.

\subsection{Evaluation of Bare Noun Disambiguation}
To measure how well our models translate BNs, we randomly select $100$ sentences and evaluate the percentage of correct BN translations for each model. The $100$ sentences selected contained $236$ occurrences of BNs. There were $9$ indefinite plural, $46$ indefinite singular, $35$ definite plural, $103$ definite singular and $43$ generic examples in this randomly selected set. We perform a detailed error analysis on the $100$ sentences for each case of incorrect BN translation. 

\begin{figure}[H]
\begin{subfigure}{\columnwidth} 
\includegraphics[width=\columnwidth]{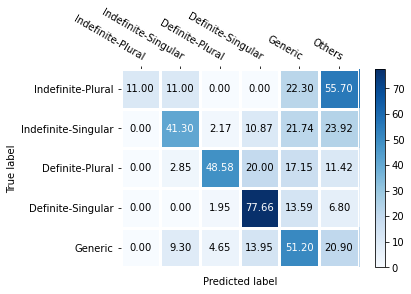}    
\caption{SMT }    
\end{subfigure}    
\hspace{\fill}  
\begin{subfigure}{\columnwidth} 
\includegraphics[width=\columnwidth]{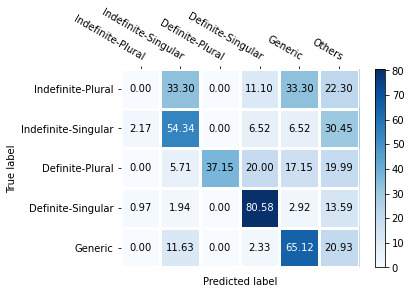}    
\caption{BiLSTM}    
\end{subfigure}
\hspace{\fill}
\begin{subfigure}{\columnwidth} 
\includegraphics[width=\columnwidth]{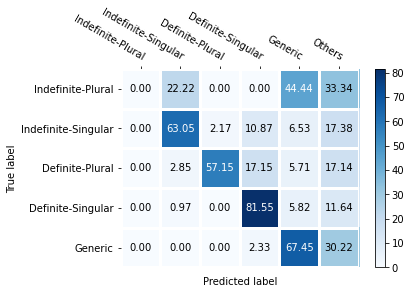}  
\caption{Transformer}    
\end{subfigure}
\caption{Confusion matrices showing the disambiguation patterns of the SMT, BiLSTM, and Transformer Models.} \label{fig:confmatrix}
\end{figure}

We assemble the type of errors found in the translation output into a confusion matrix that captures the behaviour of the $3$ models in Figure ~\ref{fig:confmatrix}. Each row in the confusion matrix represents one of our $5$ noun categories and each row adds up to $100\%$.The dark blue coloured boxes represent the correctly translated values in percentages. The light blue boxes are the incorrectly translated categories that fell within one of the $5\%$ categories. The column "others" contain incorrectly translated choices outside the $5$ categories. We discuss these type of errors in detail in ~\ref{sec:error_analysis}.

The results show that the transformer model outperforms the SMT and BILSTM models for all categories except the indefinite plural category where the SMT outperforms the Transformer model. The BiLSTM also outperforms the SMT for $3$ categories. 

\textbf{How the models handle indefinites.}
In the context of indefinites, we found that all $3$ models achieved low accuracy scores. We assume this to be the case because fewer cases of indefinites are often reported in languages like Y{o}r\`{u}b\'{a} that lack overt definite and indefinite markers and this will be represented in the data we used for training. Languages like Y{o}r\`{u}b\'{a} that do not have overt definite and indefinite determiners are ambiguous between definite and indefinite readings. Indefinites are blocked when the common ground establishes the uniqueness, or familiarity of the set denoted by the noun ~\cite{dayal2019determiners}. It is expected that indefinites can occur everywhere else. However, in texts such as these, context introduces either familiarity or uniqueness therefore reducing the occurrence of indefinites. 

\textbf{How the models handle definites.}
There were $138$ BN instances that translated as definites in the set we evaluated. This is due to the aforementioned system of disambiguation that assigns BNs to the definite class if they occur in an environment of uniqueness or familiarity. There are also certain words that have an inherent unique meaning due to cultural beliefs of Y{o}r\`{u}b\'{a} people. Titles such as \textsubdot{o}ba \textit{king}, ol\'{u}wa \textit{lord} and many more often have a definite translation because it is culturally believed for instance that only one king can rule a domain. The models show an improvement in translating definites when compared to indefinites. 

\textbf{How the models handle generics.}
The Transformer model outperforms the BiLSTM and SMT models for this category while the BiLSTM outperforms the SMT.

\subsection{Error Analysis}\label{sec:error_analysis}
We focus on other errors, that is those errors presented in the "others" category/column in Figure~\ref{fig:confmatrix}. For this class of errors, the wrongly translated BN did not translate into one of definite singular or plural, indefinite singular or plural, generic or BN category. We therefore evaluate the errors found in the sentences. For the sentences evaluated, we focus only on errors that involve nouns and determiners and ignore errors relating to other classes of words. This means that if a sentence had errors with, for example, verbs or adjectives, we ignored these errors. We categorize the errors we found and provide examples errors for each category in the SMT and NMT models in Table~\ref{tab:error-analysis}. We \textbf{bold} face relevant words and phrases occurring in the gold data that had errors in the model output. 
\\

\begin{table*}[!ht]

\centering
\begin{adjustbox}{max width=\textwidth}
\renewcommand{\arraystretch}{1.15}
{
        \begin{tabular}{>{}clll}
        \toprule

      \textbf { \small Error}    &\textbf { \small Model }   &\textbf {\small Gold } & \textbf{ \small Model Output} \\    \toprule

 \multicolumn{1}{c}{}  & \multirow{1}{*}{\textbf{SMT}   } & with each bull prepare a grain offering of three-tenths of \textbf{an ephah} of fine flour & with each bull prepare a drink offering with three-tenths of fine flour  \\ 
 \multicolumn{1}{c}{}  &     &who shut up the sea behind doors when it burst forth from \textbf{the womb}, &   or who shut the doors of the sea, when he flow back as if he had the,  \\ 
\cline{2-4}

 \multicolumn{1}{c}{}  & \multirow{1}{*}{\textbf{BiLSTM}   } & as in \textbf{the days} when you came out of egypt, i will show them my wonders.     &   as soon as he came out of egypt , i will show him all the wonders he has done .\\ 
 \multicolumn{1}{c}{}  &     &and he inserted the poles into \textbf{the rings} on the sides of the ark to carry it.&   then he put the poles on each side of the ark to put it on the chest .\\ 
\cline{2-4}

 \multicolumn{1}{c}{}  &   \multirow{1}{*}{\textbf{Transf}   }   &now hiram had sent to \textbf{the king} 120 talents of gold.&      and hiram sent him 120 talents of gold \\ 
 \multicolumn{1}{c}{\multirow{-6}{*}{\rotatebox[origin=c]{90}{\textbf{\small Missing word }}}}  &    & each day one ox, six choice sheep and \textbf{some poultry} were prepared for me,...&  each day an ox, a choice sheep and six days they provide for me... \\ 

\cline{2-4}

    \toprule 
 
 \multicolumn{1}{c}{}  & \multirow{1}{*}{\textbf{SMT}   } & everyone who quotes \textbf{proverbs} will quote this \textbf{proverb} about you: like mother, like daughter &  all those who were powe, it will this powe you: as mothers, so his son of woman.\\ 
 \multicolumn{1}{c}{}  &     & ...he who had received \textbf{the promises} was about to sacrifice his one and only son, &      ...he who receive \textbf{the promised} almost ready to take your son into one sacrifice.\\ 
\cline{2-4}

 \multicolumn{1}{c}{}  & \multirow{1}{*}{\textbf{BiLSTM}   } & the lot settles \textbf{disputes} and keeps strong opponents apart. &      lot lays up the battle and makes up two unchange as from each other . \\ 
 \multicolumn{1}{c}{}  &     & he spoke, and there came swarms of \textbf{flies}, and gnats throughout their country. & he spoke, and the kind of reitution came and became gnats in their land \\ 
\cline{2-4}
    
 \multicolumn{1}{c}{}  &   \multirow{1}{*}{\textbf{Transf}   }   & the lot settles \textbf{disputes} and keeps strong opponents apart. &  the snow finish quarreling and two oppose each other\\ 
 \multicolumn{1}{c}{\multirow{-6}{*}{\rotatebox[origin=c]{90}{\textbf{\small Wrong word / spelling}}}}  &    & with \textbf{an opening} in \textbf{the center} of \textbf{the robe}... & with \textbf{the holes} among \textbf{the belt}...\\ 
\cline{2-4}
      
    \toprule 
 \multicolumn{1}{c}{}  & \multirow{1}{*}{\textbf{SMT}   } &this is \textbf{a decree} for israel, an ordinance of the god of jacob. &      this is a lasting ordinance for israel, and the law of the god of jacob.  \\ 
 \multicolumn{1}{c}{}  &     &\textbf{the simple} inherit folly, but \textbf{the prudent }are crowned with knowledge. & a simple inherit folly but to be wise in the crown of knowledge . \\ 
\cline{2-4}

 \multicolumn{1}{c}{}  & \multirow{1}{*}{\textbf{BiLSTM}   } & as for \textbf{the donkeys} you lost three days ago, do not worry about them; they have been found... &    for \textbf{a donkeys} he was three days ago, not terrified by them; they were found to be found \\ 
 \multicolumn{1}{c}{}  &     &  and we know that in all things god works for the good of those who love him ...&  We know that everything is in good deeds for those who love god,...\\ 
\cline{2-4}
    
 \multicolumn{1}{c}{}  &   \multirow{1}{*}{\textbf{Transf}   }   &  there will be \textbf{a highway} for the remnant of his people ...&  good \textbf{ways} will be for his people,...\\ 
 \multicolumn{1}{c}{\multirow{-6}{*}{\rotatebox[origin=c]{90}{\textbf{\small Grammaticality }}}}  &    & overlay the frames with gold and make gold rings to hold \textbf{the crossbars.} &    overlay the frames with gold and make gold rings so they can be \textbf{crossbars}... \\ 

\cline{2-4}
    \toprule 
 
 \multicolumn{1}{c}{}  & \multirow{1}{*}{\textbf{SMT}   } & by faith abraham, when god tested him, offered \textbf{isaac} as a sacrifice.  &   by faith of abraham, when he was tempted to, isaac offer sacrifices, \\ 
 \multicolumn{1}{c}{}  &     & ...the next day the south wind came up, & ...the next day, forth the south wind began to blow,... \\ 
\cline{2-4}

 \multicolumn{1}{c}{}  & \multirow{1}{*}{\textbf{BiLSTM}   } & ...i will give my daughter acsah in \textbf{marriage} to the man who attacks and captures kiriath sepher. & ...i'll give my daughter acsah to a man who struck down kiriath sepher and took him \textbf{wedding}.\\ 
 \multicolumn{1}{c}{}  &     &  do not love sleep or you will grow poor; stay awake and you will have food to spare &     do not love a sleep, or you will be poor. do not sleep and have food to give you something to eat .\\ 
\cline{2-4}
    
 \multicolumn{1}{c}{}  &   \multirow{1}{*}{\textbf{Transf}   }   & about this time next year, elisha said, you will hold \textit{\textbf{a son} in your arms}. &     elisha said, this time is coming, you will take \textit{your hand for your son}. \\ 
 \multicolumn{1}{c}{\multirow{-6}{*}{\rotatebox[origin=c]{90}{\textbf{\small Wrong word order}}}}  &    & ...israel served to get a wife, and to pay for her he tended sheep.& ...israel worshiped as a wife and took care of the meat to pay \textbf{the bride} for money.\\ 

\cline{2-4}
    \toprule 
\end{tabular}}
\end{adjustbox}
\caption{Example of errors for the three models under each error category we described. The English Gold is the NIV translation for the Y{o}r\`{u}b\'{a} Source }\label{tab:error-analysis}
\end{table*}
\noindent \textbf{Missing word.} Outputs with missing nouns or determiners are classified under this category.  

\noindent \textbf{Wrong word or spelling.} Wrong use of determiners or incorrect nouns belong to this category. We also classify wrong spellings, poorly inflected forms of the noun and unknown words under this category.

\noindent \textbf{Grammaticality.} We categorize both syntactic and semantic errors here. Wrong tense or aspect, incorrect number, poor punctuation, and lack of coherence are categorized under this class.

\noindent \textbf{Wrong word order or category.} If the order of the noun and determiner is wrong, even if the determiner and nouns are correct, we classify this as an error. We also include instances where the order of the noun and determiner occurs inappropriately, either before the verb or any other category. We found that some instances require both word level and phrase level order errors. In the case of word level order errors, we can generate a correct sentence by moving individual words, independently of each other, whereas for a phrase level order error, blocks of consecutive words should be moved together to form a right translation. In addition, this category includes instances in which a different type is used instead of a noun or determiner.

In evaluating each category, we do not consider correct synonyms as errors.

\subsection{Word-Level BN Disambiguation }\label{sec:word_level_perf}
We randomly select $10$ nouns that occur as BNs in the test data and check how well the $3$ models disambiguate these words. We use the following words: \`{i}l\'{u} \textit{town}, \`{i}w\'{e} \textit{book}, il\'{e} \textit{house}, \textsubdot{o}k\`{u}nrin \textit{male},  ob\`{i}nrin \textit{female},   a\textsubdot{s}\textsubdot{o} \textit{clothing}, il\textsubdot{\`{e}} \textit{land}, \textsubdot{o}ba  \textit{king}, \`{a}l\`{u}f\'{a}\`{a} \textit{priest} and baba \textit{father}. 

We then calculate the percentage of correct translations of these BNs in the test set. We use the English gold data to determine the correct disambiguation and compare each instance of the BNs with the corresponding occurrence in the gold data. 
\hspace{\fill} 
\begin{figure}[!t]
\centering

  \includegraphics[width=6cm]{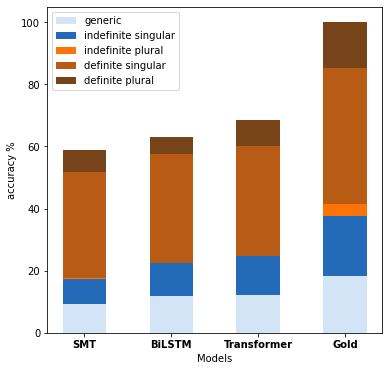}
  \caption{Distribution of disambiguation patterns in our models. Our gold data has $100\%$ in disambiguation.}
\label{fig:dis}
\end{figure}

Our analysis in Figure~\ref{fig:dis}  shows that the transformer model performs better in disambiguating the BNs selected. The transformer model achieves \textbf{67.85\%} accuracy while the BiLSTM model achieves an accuracy of \textbf{63.83\%}. The SMT model, on the other hand, achieves an accuracy of \textbf{59.12\%}. 
\section{Conclusion}\label{sec:conc}
In this work, we showed how SMT, BiLSTM, and transformer models translate BNs in Y{o}r\`{u}b\'{a}, a resource scarce language. We compared ability of SMT and NMT models to correctly translate BNs into various categories referenced in the syntax literature of Y{o}r\`{u}b\'{a}. We measured the performance of the MT models and the output using BLEU scores, and by counting the percentage of correctly disambiguated BNs compared with incorrectly disambiguated BNs. We found a positive correlation between disambiguation accuracy, and BLEU scores as well as a positive correlation between number of occurrences of a category and the accuracy in translation.

We also found the transformer outperforming the SMT and BiLSTM models in correctly translating BNs and found that all $3$ models best performed in translating a BN in Y{o}r\`{u}b\'{a} into an definite singular in English. This finding corroborates research that predicts that languages which lack overt definite and indefinite markers have larger cases of definites, and findings within the MT community that MT models improve with more data. We also analyzed the type of errors our systems produce. We identified cases of missing words, wrong word or spellings, grammaticality issues, and word-order errors. We found that even when certain BNs have been correctly categorized by the models, the models still had semantic, and or syntactic errors. 

To further probe the capacity of SMT and NMT models in disambiguating BNs, we can improve the SMT and NMT models with back-translation, cross-lingual word embeddings, larger training data, among other approaches and perform human based evaluations on the entire quality of the MT output. Standard test sets can also be developed to aid automatic comparison of human evaluations and machine based evaluations. In addition, the Y{o}r\`{u}b\'{a} data we used for this work is a translated document; a translation from English to Y{o}r\`{u}b\'{a} and it will be interesting to use a text originally written in Y{o}r\`{u}b\'{a} but translated to English for this experiment. 
\section*{Acknowledgements}
\small{We gratefully acknowledge support from the Natural Sciences and Engineering Research Council of Canada (NSERC), the Social Sciences Research Council of Canada (SSHRC), Compute Canada (\url{www.computecanada.ca}), and UBC ARC--Sockeye (\url{https://doi.org/10.14288/SOCKEYE}).}

\bibliography{anthology,eacl2021}
\bibliographystyle{acl_natbib}
\end{document}